\pdfoutput=1

\documentclass[11pt]{article}

\usepackage[final]{acl}

\usepackage{times}
\usepackage{latexsym}

\usepackage[T1]{fontenc}

\usepackage[utf8]{inputenc}

\usepackage{microtype}

\usepackage{inconsolata}

\usepackage{graphicx}

\usepackage{color}
\usepackage{CJKutf8}
\usepackage{amsfonts}
\usepackage{multirow}
\usepackage{array}
\usepackage{booktabs}
\usepackage{subcaption}
\usepackage{colortbl, booktabs}
\usepackage{enumitem}
\usepackage{float}
\usepackage{makecell}

\title{The Effects of Data Augmentation on Confidence Estimation for LLMs}

\author{
 \textbf{Rui Wang\textsuperscript{1,2}},
 \textbf{Renyu Zhu\textsuperscript{3}},
 \textbf{Minmin Lin\textsuperscript{3}},\\
 \textbf{Runze Wu\textsuperscript{3}},
 \textbf{Tangjie Lv\textsuperscript{3}},
 \textbf{Changjie Fan\textsuperscript{3}},
 \textbf{Haobo Wang\textsuperscript{1,2}\thanks{\, \,Corresponding authors.}}
\\
 \textsuperscript{1}School of Software Technology, Zhejiang University\\
 \textsuperscript{2} Hangzhou High-Tech Zone (Binjiang) Institute of Blockchain and Data Security \\
 \textsuperscript{3}NetEase Fuxi AI Lab
\\
{\small \texttt{\{22351017,wanghaobo\}@zju.edu.cn} } \\\ {\small \texttt{\{zhurenyu,linminmin01,wurunze1,hzlvtangjie,fanchangjie\}@corp.netease.com}}
}
\begin{document}
\maketitle
\begin{abstract}

Confidence estimation is crucial for reflecting the reliability of large language models (LLMs), particularly in the widely used closed-source models. 
Utilizing data augmentation for confidence estimation is viable, but discussions focus on specific augmentation techniques, limiting its potential. We study the impact of different data augmentation methods on confidence estimation.
Our findings indicate that data augmentation strategies can achieve better performance and mitigate the impact of overconfidence. We investigate the influential factors related to this and discover that, while preserving semantic information, greater data diversity enhances the effectiveness of augmentation. Furthermore, the impact of different augmentation strategies varies across different range of application. Considering parameter transferability and usability, the random combination of augmentations is a promising choice. 

\end{abstract}

\section{Introduction}

Although LLMs~\cite{palm,gpt4} exhibit remarkable capabilities in generalization across various natural language processing (NLP) tasks, their tendency to generate non-factual responses~\cite{h1} raises significant concerns. It is essential to assess the reliability of their generalization results. As black-box LLMs become more prevalent, accessing their internal information is challenging, which increase the difficulty of evaluate the reliability.
Confidence estimation~\cite{just,b2,ZhangHSGPYZQ24} has emerged as a popular solution, facilitating risk assessment and error checking.

There is a wide variety of methods~\cite{w1,w2,w3} for confidence in white-box LLMs, but there are few types of methods for black-box LLMs, where the common methods are Bayesian methods~\cite{b2,b1}, particularly in sampling strategy~\cite{SiGYWWBW23}. It inputs the original text into the LLM and samples for predictions.
In addition to that, some researchers~\cite{aug_ner,b2,spuq} using data augmentation for confidence estimation. Arguably, it is a promising strategy for LLM's confidence estimation, especially for black-box ones with inaccessible parameters.
However, they either operate only on white-box small models or engage in very limited discussions regarding augmentation methods, relying on weak augmentation or uncontrollable augmentation. Therefore, we hope to conduct a comprehensive evaluation of confidence estimation based on data augmentation to verify its impact. 

In this study, we explore common augmentation strategies~\cite{eda,SennrichHB16} and discuss typical automated augmentation methods~\cite{CubukZSL20,ren2021taa} to validate the effectiveness of combining augmentation techniques. In particular, we investigate the following research questions:
\begin{itemize}[itemsep=0pt,topsep=0pt,parsep=3pt]
    \item \textbf{Q1}: What impact does data augmentation have on confidence estimation?
    \item \textbf{Q2}: What factor contributes to this impact?
    \item \textbf{Q3}: Is the range of applications for different data augmentation techniques consistent?
\end{itemize}

Comprehensive experiments on benchmark datasets show that data augmentation is effective for confidence estimation and mitigates the impact of LLMs' overconfidence. The best data augmentation method reduces the average ECE across three models from 11.50\% to 5.97\% in GSM8K.
We perform further analysis of the experimental results to explore which factors impact the results of data augmentation strategies on confidence estimation. Our findings indicate that data diversity and semantic consistency are key. While maintaining semantic information, higher data diversity leads to improved confidence estimation. Notably, RandAugment~\cite{CubukZSL20} demonstrates better performance and exhibits significant potential for cross-model transfer.
Moreover, our study reveals that different data augmentation methods have different ranges of applicability. A mild augmentation strategy is more appropriate for mathematical data about complex logical reasoning. Therefore, when addressing downstream tasks that are unknown, it is advisable to use RandAugment to compute confidence. We believe this will contribute to enhancing the reliability of generalization in LLMs.

\section{Related Works}
\textbf{Confidence Estimation for Black-Box LLMs. }
The confidence estimation of black-box models is divided into four categories by common methods. Single deterministic methods~\cite{LinHE22,just} rely on verbal descriptions, which are hard to align with the model's internals~\cite{KumarMUKE24}. Ensemble methods~\cite{ZhangLDMS23} based on multiple LLMs diverge from our focus on a single model, so we won't explore this. While Bayesian methods~\cite{b2,ZhangHSGPYZQ24} are common, LLMs' overconfidence undermines their effectiveness. In the methods that use augmentation, the weak augmentation~\cite{spuq} struggles to enhance text diversity. Paraphrasing method~\cite{b2} lacks controllability. 
This makes it difficult to mitigate overconfidence.

\noindent\textbf{Data Augmentation.}
Data augmentation~\cite{eda,FengGWCVMH21} is widely used in NLP, primarily includes insertion, deletion, replacement, and swapping. Back-translation~\cite{SennrichHB16} is useful in low-resource domains. With the rise of LLMs, paraphrasing method~\cite{PiedboeufL23} using LLMs rewrite the text, but ensuring quality can be difficult and expensive. Besides, automated augmentation has gained attention. Text AutoAugment~\cite{ren2021taa} uses automated augmentation in NLP, while RandAugment~\cite{CubukZSL20} reduces retrieval costs. In subsequent experiments, we examine traditional and automated augmentation, revealing their potential in confidence estimation.

\begin{figure}[!t]
  \centering
  \vspace{-0.1cm}
  \includegraphics[width=0.92\linewidth]{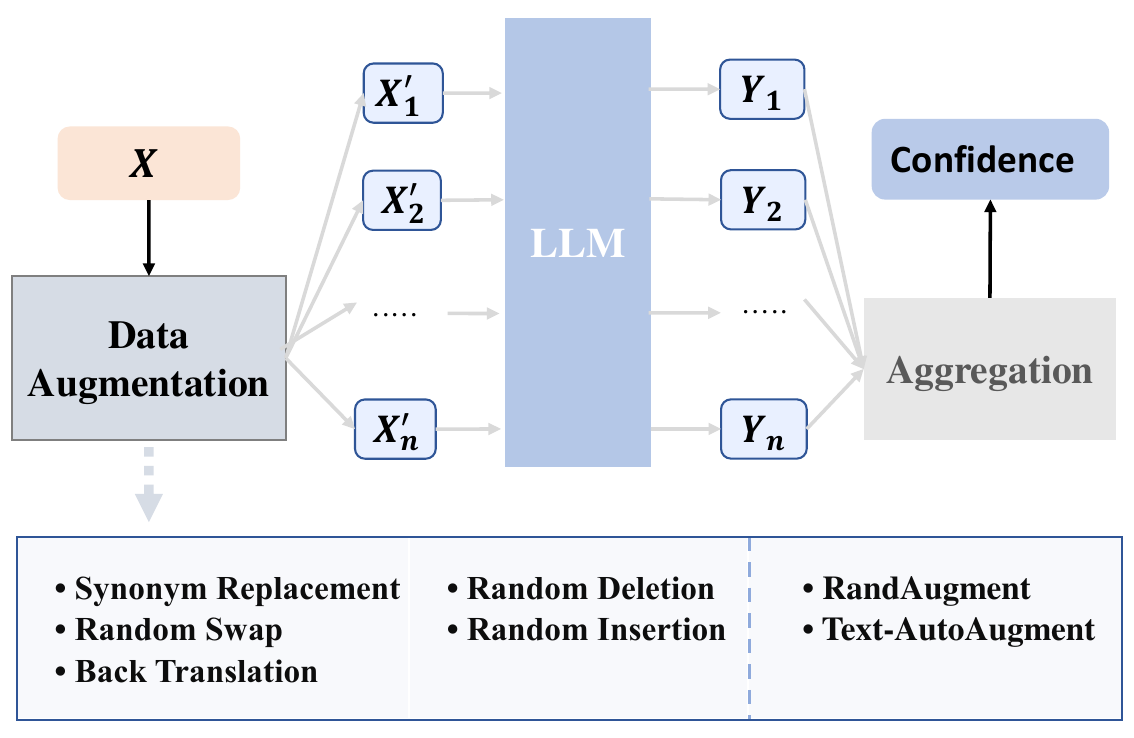}
   \caption{
   Overview of a confidence estimation framework utilizing data augmentation. Each sample $X$ is subjected to $n$ augmentations, yielding a more diverse array of augmented instances $X'$. The LLM performs predictions $Y$ on these augmented samples, and confidence is derived through aggregation.
   }
   \label{fig:fig1}
\end{figure}

\section{Methodology}

\begin{table*}[tbhp]\centering\small
\linespread{1}\selectfont
\setlength{\tabcolsep}{2.5pt} 
\begin{tabular}{lcccccccccc}
\toprule
\multirow{2}{*}{Method} & \multicolumn{3}{c}{StrategyQA}                                                  & \multicolumn{3}{c}{Professional Law}                                      & \multicolumn{3}{c}{GSM8K}  &\multirow{2}{*}{AVG}                                                 \\
\cmidrule(lr){2-10}
                        & Qwen2       & \multicolumn{1}{c}{Llama3}      & \multicolumn{1}{c}{Gemma2}      & Qwen2       & \multicolumn{1}{c}{Llama3} & \multicolumn{1}{c}{Gemma2}      & Qwen2       & \multicolumn{1}{c}{Llama3}     & \multicolumn{1}{c}{Gemma2} \\
\midrule
Sampling                & 27.00 & 28.25                    & 22.14                    & 42.04 & 39.32                    & 33.22                    & 12.94 & 12.14                    & 9.41                    & 25.16\\
Paraphrase            & 33.57 & 35.33                    & 39.42                    & 48.46 & 46.22                   & 45.54                    & 46.13 & 61.44                    & 67.46    & 47.06 \\
\midrule
Synonym Replacement & 23.72 & \underline{26.01}                    & 20.21                    & 38.98 & \underline{36.41}                    & 30.86                    & \textbf{5.38} & 8.56                    & \textbf{4.72}                    & 21.65 \\
Random Swap           & 23.81  & 29.38                    & 20.76                    & \underline{35.58} & 36.96                    & 30.39                    & 6.75 & 10.48                    & 25.09                  & 24.36  \\
Random Deletion        & \underline{20.46}  & \textbf{20.81}                    & \textbf{16.28}                    & 37.65 & 36.56                    & \underline{29.73}                     & 8.74 & \underline{7.81}                    & 13.04                    & \underline{21.23} \\
Random Insertion       & 26.66 & 28.37                    & 20.61                    & 38.41 & 37.29                    & 32.30                    & 13.44 & 14.36                     & 11.15                                   & 24.73 \\
Back-Translation             & 24.13 & 27.94                     & 22.20                    & 46.67 & 44.59                    & 38.03                    & 8.05 & 8.36                    & \underline{6.14}                    & 25.12 \\
\midrule
RandAugment               & 20.55 & 26.73                    & \underline{16.46}                    & \textbf{33.13} & 36.85                    & \textbf{26.92}                    & \underline{6.55} & \textbf{7.17}                     & 6.15                  & \textbf{20.05} \\
TAA                     & \textbf{18.98} & 27.38                    & 20.61                    & 38.17 & \textbf{34.96}                    & 31.05                    & 7.65 & 19.27                    & 15.38                         & 23.72 \\
\bottomrule
\end{tabular}
\caption{Confidence estimation of 3 models (metrics are given by $×10^2$).
The evaluation metric is ECE(↓). The best results are marked in bold and the second-best marked in underline. 
}\label{tab:tab1}
\end{table*}

\begin{table*}[tbhp]\centering\small
\linespread{1}\selectfont
\setlength{\tabcolsep}{3pt} 
\begin{tabular}{lccccccc}
\toprule
\multirow{2}{*}{Method} & \multicolumn{2}{c}{StrategyQA}                                                  & \multicolumn{2}{c}{Professional Law}                                      & \multicolumn{2}{c}{GSM8K}    &\multirow{2}{*}{AVG}                                             \\
\cmidrule(lr){2-7}
                             & \multicolumn{1}{c}{GPT-3.5}      & GPT-4o-mini           &GPT-3.5 & GPT-4o-mini          & GPT-3.5     & GPT-4o-mini \\
\midrule
Sampling & \underline{18.62} & 18.45 & 30.77 & 29.36 & 8.94  & 3.83  & 18.33\\
\midrule
Synonym Replacement & 19.59 & \underline{16.05} & 31.92 & 28.54 & 7.33 & \underline{3.81} & 17.87\\
Random Deletion & \textbf{16.68} & \textbf{10.86}  & \underline{27.86} & \underline{26.72} & \textbf{4.84}  & \textbf{3.75}   & \textbf{15.12} \\
\midrule
RandAugment     & 18.68  & 16.73 & \textbf{25.42} & \textbf{22.97} & \underline{7.11}  &7.02    & \underline{16.32} \\
\bottomrule
\end{tabular}
\caption{Further experiments about the best method from each category for the closed-source model (metrics are given by $×10^2$). The evaluation metric is ECE(↓). 
}\label{tab:tab2}
\end{table*}

\subsection{Confidence Estimation}
We utilize data augmentation strategies to compute confidence.
Specifically, given a sample $X$, it undergoes augmentation $n$ times to generate $n$ augmented samples ${\cal S}=[X'_1, X'_2, ..., X'_n]$, where $X'_i=s_i(X)$ and $s_i(\cdot)$ is the $i$-th data augmentation for the sample. 
These augmented samples ${\cal S}$ are input into the LLM $f$ to produce predictions ${\cal A}=[Y_1, Y_2, ..., Y_n]$, i.e., $Y_i=f(X'_i)$.
Subsequently, a consistency strategy~\cite{SiGYWWBW23} is applied to aggregate the answers. The confidence $C$ about $X$ is the consistency of answer $\bar{Y}$ in ${\cal A}$,
\begin{equation}
C=\frac{1}{n}\sum_{i=1}^{n}{\mathbb{I}\{Y_i=\bar{Y}\}},
\label{eq1}
\end{equation}
where $\bar{Y}$ is the most frequently predicted answer and $\mathbb{I}$ is indicator function.

\subsection{Data Augmentation Methods}
The selection of augmentation strategy is important in our confidence computation. We believe that an effective augmentation strategy should be controllable and simple. Consequently, we choose the straightforward method from the four most common categories of traditional augmentation techniques: \textbf{synonym replacement}, \textbf{random swap}, \textbf{random deletion}, and \textbf{random insertion}.
Each method enables control over the degree of augmentation by adjusting the magnitude. Besides, we discuss a popular data augmentation technique, \textbf{back-translation}~\cite{SennrichHB16}.

To examine the synergistic effects of traditional augmentation strategies, we discuss two automated augmentation methods: \textbf{RandAugment}~\cite{CubukZSL20} and \textbf{Text-AutoAugment (TAA)}~\cite{ren2021taa}. RandAugment randomly combines traditional augmentation strategies, while TAA applies a fixed set of augmentation to each sample. They need to find the best parameters in the validation set and use them for the test set. More details are described in the Appendix.

\section{Experiments}

\subsection{Experimental Setup}

To discuss the impact of data augmentation on confidence estimation, we conduct experiments using Llama-3-8b~\cite{llama3modelcard}, Gemma-2-9b~\cite{gemma2}, and Qwen-2-7b~\cite{qwen2}. We use GPT-3.5, and GPT-4o-mini for further experimental verification. The implementation details are in the Appendix~\ref{detail}.

\textbf{Datasets.} We evaluate the quality of confidence estimates across three datasets: 1) \textit{StrategyQA}~\cite{SQA} 
from BigBench~\cite{bigbench}
, which is about commonsense reasoning
; 2) \textit{Professional Law} (Prf-Law) from MMLU~\cite{MMLU}, which is about professional knowledge
; 3) \textit{GSM8K}~\cite{gsm8k} is about math word problems. 

\textbf{Baselines.} We select representative strategies from two categories. \textbf{The baselines utilize Eq.(\ref{eq1}) for aggregation}.
For the Bayesian method, we use \textit{sampling strategy}~\cite{SiGYWWBW23} that inputs the original text into the LLM, with responses sampled at a high temperature. Besides, we use a \textit{paraphrasing-based method}, which paraphrases questions using LLMs~\cite{b2}.

\textbf{Evaluation Metric.} To assess the alignment between confidence and accuracy, we introduce Expected Calibration Error (ECE)~\cite{ece}. It calculates the difference between the average confidence and the accuracy of the model.

\subsection{Results}

\paragraph{Answer for Q1: Data augmentation is beneficial for confidence estimation, which mitigates the impact of LLMs' overconfidence.}
As shown in Table~\ref{tab:tab1}, the average ECE of all augmentation strategies outperforms the sampling method. The top three performing augmentation strategies are \textit{RandAugment, Random Deletion, and Synonym Replacement}. In GSM8K, Synonym Replacement reduces the average ECE across three models from 11.50\% to 5.97\%. Additionally, there are decreases of 6.61\% and 5.89\% in StrategyQA and Professional Law, respectively. In Table~\ref{tab:tab2}, the average ECE of data augmentation methods still demonstrates performance advantages.

\begin{table}[t]\centering\small
\linespread{1}\selectfont
\setlength{\tabcolsep}{2.5pt} 
\begin{tabular}{lccc}
\toprule
Method & StrategyQA                                                  & Prf-Law &GSM8K  \\
\midrule
Sampling     & 0.89    & 0.84    & 0.44  \\
\midrule
Synonym Replacement     & 0.85  & 0.79    & 0.36  \\
Random Deletion     & 0.79   & 0.76 & 0.35   \\
\midrule
RandAugment     & 0.79    & 0.72   & 0.34 \\

\bottomrule
\end{tabular}
\caption{Mean confidence for Qwen2, Llama3, and Gemma2. The table presents the average confidence (↓) of incorrectly predicted samples. 
}\label{tab:tab3}
\end{table}

The hallucinations of LLMs often lead to overconfidence in incorrect predictions. We compare the confidence of three optimal augmentation strategies on incorrect samples to further investigate the impact of data augmentation.
Ideally, the confidence for incorrect samples should be 0. However, as shown in Table~\ref{tab:tab3}, the sampling method exhibits high confidence, particularly in StrategyQA and Professional Law, where the confidence reaches up to 0.89 and 0.84. Based on data augmentation strategies, the overconfidence of LLMs is alleviated, and the confidence is reduced to 0.79 and 0.72. It shows that data augmentation holds significant potential for confidence estimation.

\paragraph{Answer for Q2: Data diversity and semantic consistency are important. While maintaining semantic integrity, the more diverse the samples are, the better the outcomes will be.}
Based on the analysis of Table~\ref{tab:tab1}, the degradation of semantic information, such as random swap, results in poorer outcomes. Similarly, random insertion introduces noisy information that can affect the judgments of LLMs. Moreover, over-focusing on semantic consistency and ignoring sample diversity can also be detrimental to confidence estimation, as seen with back-translation techniques. We speculate that the effectiveness of random deletion is from the LLM's inherent ability to infer missing tokens, allowing it to extract meaningful information from diverse augmented samples. RandAugment ranks highest in average ECE in Table~\ref{tab:tab1} because its random combinations of individual strategies enhance sample diversity. In contrast, the fixed augmentation combinations used in TAA result in lower diversity compared to RandAugment, which is why TAA performs worse than RandAugment.

We conduct a further analysis of RandAugment and find that it demonstrates \textbf{cross-model adaptability}. Specifically, the augmentation combinations learned by a model can be transferred to other models. In Tables~\ref{tab:tab4}, we apply the best parameters from Qwen2 to Llama3 and Gemma2, resulting in performance improvements in most cases compared to the sampling method. In StrategyQA, the best parameters from Llama3 and Qwen2 are the same, so they achieve the same results.

\begin{table}[!t]\centering\small
\setlength{\tabcolsep}{2.5pt} 
\begin{tabular}{llccc}
\toprule
Model    & Method          & StrategyQA & Prf-Law & GSM8K \\
\midrule
\multirow{3}{*}{Llama3} &Sampling    & 28.25 & 39.32 & 12.14  \\
\cmidrule(lr){2-5}
&RandAugment    &26.73  & 36.85 & 7.17 \\
&\cellcolor{gray!30} + Parameters     & \cellcolor{gray!30}26.73 &\cellcolor{gray!30} 36.27  &\cellcolor{gray!30} 15.61  \\
\midrule
\multirow{3}{*}{Gemma2}  & Sampling   & 22.14    & 33.22          & 9.41      \\
\cmidrule(lr){2-5}
&RandAugment & 16.46    & 26.92          & 6.15       \\
&\cellcolor{gray!30} + Parameters     & \cellcolor{gray!30}   19.62  &\cellcolor{gray!30}  24.97 &\cellcolor{gray!30} 9.33    \\
\bottomrule
\end{tabular}
\caption{Apply the RandAugment parameters from the smallest model, Qwen2-7b, to the larger models, Llama-3-8b and Gemma-2-9b, as indicated in grey. The evaluation metric is ECE(↓). Metrics are given by $×10^2$.
}\label{tab:tab4}
\end{table}

\paragraph{Answer for Q3: Different data augmentation techniques have different ranges of applications.
Moderate strategies are recommended only for math data that requires complex reasoning.}
In Table~\ref{tab:tab1} and Table~\ref{tab:tab2}, 
RandAugment and random deletion, which can introduce more diverse samples, generally achieve favorable outcomes across a range of datasets. However, relatively mild augmentation strategies often demonstrate significant dataset-specific biases.
Back-translation perform better on GSM8K compared to StrategyQA and Professional Law. While back-translation results underperforms on Professional Law, it reduces the ECE on GSM8K from 11.50\% to 7.52\%. For a math dataset that prioritizes logical reasoning, the complexity of the questions and the requirement for logical reasoning make the dataset more sensitive to sample diversity. Thus, milder augmentation strategies remain effective. 
However, for StrategyQA and Professional Law, the strong common-sense reasoning and extensive domain knowledge in LLMs render overly cautious augmentation strategies ineffective. 

\section{Discussion}
LLMs are sensitive to prompts~\cite{prompt}, making responses from diverse views align better with their cognition, leading to the effectiveness of data augmentation in confidence estimation. Given the different applicability of augmentation strategies, we recommend using RandAugment for confidence calculation in unknown downstream tasks. This is due to its cross-model adaptability and usability. We are confident that the confidence estimation based on data augmentation will beneficial for the reliability of LLM generalization, supporting the development of safer and reliable NLP systems.

\section*{Limitation}
While we conduct experiments and validate the effectiveness of data augmentation in confidence estimation, there is a limitation that should be acknowledged.
Regarding the data augmentation strategies, due to resource limitations, we don't discuss all possible augmentation techniques; instead, we only consider five typical traditional augmentation strategies and two basic automatic augmentation strategies. We acknowledge the possibility that some augmentation methods that we don't discuss may be more suitable in certain contexts than others that we have considered.

\section*{Ethical Considerations}
In this study, we utilized existing datasets that have already addressed ethical considerations. Additionally, the data augmentation methods employed are safe and reliable, making it unlikely that toxic sentences will be generated. This has been validated by numerous previous studies. Additionally, our manual reviews did not reveal any issues.

\bibliography{custom}

\appendix
=

\section{Additional Experimental Results}

\paragraph{Transferability across datasets.}
We discuss the adaptability of RandAugment across different datasets. It is evident that, in most cases, the parameters obtained from other datasets maintain relatively stable performance on the target dataset. This adaptability is particularly noteworthy when the target dataset is StrategyQA or Professional Law, demonstrating strong cross-dataset compatibility.

\begin{table}[H]\centering\small
\setlength{\tabcolsep}{2pt}
\begin{tabular}{ccccc}
\toprule
\multicolumn{1}{l}{Target Dataset}    & Source Dataset       & Qwen2   & Llama3  & Gemma2  \\
\midrule
\multirow{3}{*}{StrategyQA} & StrategyQA & 20.546 & 26.726 & 16.463 \\
                            & Prf-Law    & 20.813 & 27.467 & 18.713 \\
                            & GSM8K      & 21.564 & 26.167 & 16.463 \\
\midrule
\multirow{3}{*}{Prf-Law}    & StrategyQA & 32.840  & 37.001    & 24.424 \\
                            & Prf-Law    & 33.129 & 36.852 & 26.919 \\
                            & GSM8K      & 36.587 & 34.713 & 24.424 \\
\midrule
\multirow{3}{*}{GSM8K}      & StrategyQA & 16.61  & 10.571 & 6.149 \\
                            & Prf-Law    & 9.314 & 17.750  & 32.724 \\
                            & GSM8K      & 6.547 & 7.170  & 6.149 \\
\bottomrule
\end{tabular}
\caption{Apply the parameters of source dataset to target dataset.}\label{tab:tab5}
\end{table}

\begin{table*}[t]\centering\small
\begin{tabular}{c|c}
\toprule
Dataset          & Prompt\\
\midrule
StrategyQA \& Prf-Law     & \makecell[l]{Read the question, analyze step by step, provide your answer.\\
Use the following format to answer:\\
```Explanation: [insert step-by-step analysis here]\\
Answer: [ONLY the option letter; not a complete sentence]'''\\
Only give me the reply according to this format, don't give me any other words.}\\
\midrule
GSM8K     & \makecell[l]{Read the question, analyze step by step, provide your answer.\\
Use the following format to answer:\\
```Explanation: [insert step-by-step analysis here]\\
Answer: [ONLY the number; not a complete sentence]'''\\
Only give me the reply according to this format, don't give me any other words.}\\
\bottomrule
\end{tabular}
\caption{Prompt templates for each dataset.
}\label{tab:tab6}
\end{table*}

\section{Implementation Details}\label{detail}
We follow prior work~\cite{b2} to set the augmentation times to $n=5$ and keep the sampling quantity of the baseline consistent with it. The model temperature is 0 for the paraphrasing methods, and 0.7 for the sampling-based and our method. The validation set to test set ratio is 1:1. Augmentation magnitude can be selected from \{0.1, 0.2, 0.3\}. Automated augmentation retrieval process assesses the quality of augmentation combinations based on ECE, excluding model training.  For all experiments, we run three times and report the averaged results.
\subsection{Traditional Data Augmentation}
We provide supplementary information about traditional data augmentation:
\begin{itemize}
    \item \textbf{Synonym Replacement (SR).} 
    SR~\cite{eda} randomly selects words from the sentence. It then substitutes each of these words with a randomly chosen synonym.
    \item \textbf{Random Swap (RS).} 
    RS~\cite{eda} randomly swaps the positions of two words $K$ times.
    \item \textbf{Random Deletion (RD).} 
    RD~\cite{eda} randomly deletes each word based on a probability of $p$.
    \item \textbf{Random Insertion (RI).} 
    RI~\cite{eda} finds a random synonym of a random word in the sentence that is not a stop word and inserts that synonym into a random position in the sentence $K$ times.
    \item \textbf{Back-Translation.} It~\cite{SennrichHB16} typically translates text from the original language to a second language, and then back to the original language. 
    Here, we choose French as the second language.
\end{itemize}

In the experiment, aside from back-translation, which does not require an augmentation magnitude, the remaining four augmentation strategies that need to be tested on the validation set should determine the optimal augmentation magnitude before being applied to the test set.

\subsection{Automated Augmentation}
We describe the details of the automated augmentation strategies:
\begin{itemize}
    \item \textbf{RandAugment.} 
    It is necessary to determine the optimal augmentation combination on the validation set before applying it to the test set. The key parameters are the number of augmentation transformations $N_r$ and the magnitude of augmentation $M$. For each sample, $N_r$ augmentation transformations are randomly selected from five traditional augmentation operations and then applied sequentially to the sample, each applied with a magnitude of $M$. The range of values for the number of augmentation transformations \( N_r \) is \{1, 2, 3\}. The magnitude \( M \) can take values from \{0.1, 0.2, 0.3\}. A grid search should first be conducted on the validation set to determine the optimal values for \( N_r \) and \( M \), which will then be applied to the test set.
    \item \textbf{TAA.}
    It also requires determining the optimal augmentation combination first. A augmentation combination consists of $N_{t}$ policies, with each policy $P=\{O_1, O_2, ..., O_T\}$ containing $T$ editing operations $O$. Each editing operation includes an augmentation transformation, a probability of calling a transformation, and a magnitude. Each sample requires executing all editing operations in a policy, i.e., $s=O_T\circ ...\circ O_2\circ O_1$.
\end{itemize}

In TAA, the number of policies is $N_t=4$, and the number of editing operations is $T=2$, with the top three optimal combinations retained, i.e., $3*N_t$. We set the number of iterations to 50.

\subsection{Additional Implementation Details}
To ensure that the LLM thoroughly understands the questions and accurately demonstrates its capabilities while minimizing the occurrence of hallucinations, we employ zero-shot Chain-of-Thought (CoT)~\cite{cot}.

\end{document}